%% file: main.tex
\documentclass[letterpaper, 10 pt, conference]{ieeeconf}  

\IEEEoverridecommandlockouts                              
\usepackage{xcolor}
\usepackage{comment}
\usepackage{graphics} 
\usepackage{graphicx}
\usepackage{subcaption}
\usepackage{epsfig} 
\usepackage{amsmath} 
\usepackage{amssymb}  
\usepackage{mathtools}
\usepackage{marginnote}
\usepackage{bm}
\usepackage{hyperref}
\usepackage{iitem}
\usepackage{float}

\usepackage{svg}
\usepackage{balance}
\usepackage{caption}
\include{macro}

\title{\LARGE \bf
Efficient and Versatile Quadrupedal Skating: Optimal Co-design via Reinforcement Learning and Bayesian Optimization} 

\author{ Hanwen Wang$^{*1}$, Zhenlong Fang$^{*1}$, Josiah Hanna$^1$, Xiaobin Xiong$^{1,2}$ 
\thanks{$^*$Hanwen Wang and Zhenlong Fang contributed equally in this work.}%
\thanks{$^1$Hanwen Wang, Zhenlong Fang, and Josiah Hanna are with the University of Wisconsin - Madison. 
        $^2$Xiaobin Xiong was with the University of Wisconsin - Madison, and now with the Shanghai Innovation Institute (SII).
        {Corresponding to Xiaobin Xiong: \tt\small xiaobin.xiong@sii.edu.cn}.}%
}
\begin{document}

 \newcommand{\block}[1]{\noindent{\textbf{#1}:}}
\newcommand{\fix}[1]{{\color{red} #1}}

\maketitle
\thispagestyle{empty}
\pagestyle{empty}


\begin{abstract}
In this paper, we present a hardware-control co-design approach that enables efficient and versatile roller skating on quadrupedal robots equipped with passive wheels. Passive-wheel skating reduces leg inertia and improves energy efficiency, particularly at high speeds. However, the absence of direct wheel actuation tightly couples mechanical design and control. To unlock the full potential of this modality, we formulate a bilevel optimization framework: an \emph{upper-level Bayesian Optimization} searches the mechanical design space, while a \emph{lower-level Reinforcement Learning} trains a motor control policy for each candidate design. The resulting design-policy pairs not only outperform human-engineered baselines, but also exhibit versatile behaviors such as \emph{hockey stop} (rapid braking by turning sideways to maximize friction) and \emph{self-aligning} motion (automatic reorientation to improve energy efficiency in the direction of travel), offering the first system-level study of dynamic skating motion on quadrupedal robots.
\end{abstract}

\input{sections/01_introduction}

\input{sections/02_related}

\input{sections/03_quad_skating_via_codesign}

\input{sections/04_algo_impl}

\input{sections/05_results_and_analysis}

\input{sections/06_conclusion_and_future_work}

\bibliographystyle{IEEEtran}
\bibliography{ref/design_and_robot, ref/RL, ref/control, ref/BO, ref/codesign, ref/estimation}
\end{document}

%% file: macro.tex
\newcommand{\Fig}[1]{{Fig.~#1}}
\newcommand{\Table}[1]{{Table ~#1}}

\newcommand{\Sec}[1]{{Sec.~#1}}

\newcommand{\cB}{\mathcal{B}}
\newcommand{\cW}{\mathcal{W}}

\newcommand{\cT}{\mathcal{T}}
\newcommand{\bR}{\mathbf{R}}
\newcommand{\bp}{\mathbf{p}}
\newcommand{\bv}{\mathbf{v}}
\newcommand{\bomg}{\bm{\omega}}
\newcommand{\bu}{\mathbf{u}}
\newcommand{\boldf}{\mathbf{f}}
\newcommand{\bq}{\mathbf{q}}
\newcommand{\bdq}{\dot{\mathbf{q}}}
\newcommand{\btau}{\bm{\tau}}

\newcommand{\ba}{\mathbf{a}}
\newcommand{\bd}{\mathbf{d}}
\newcommand{\bxi}{\bm{\xi}}
\newcommand{\LtwoNormSquare}[1]{\left\lVert #1 \right\rVert_2^2}
\newcommand{\LtwoNorm}[1]{\left\lVert #1 \right\rVert_2}
\newcommand{\err}{\text{err}}
\newcommand{\txtnext}{\text{next}}
\newcommand{\cmd}{\text{cmd}}

%% file: sections/01_introduction.tex
\section{Introduction}

Legged robots have made rapid progress in agility and robustness, enabling increasingly capable real-world deployment~\cite{raibert2008bigdog,hwangbo2019learning,di2018dynamic}. However, compared to wheeled platforms, legged locomotion still lags in speed and energy efficiency~\cite{kim2006isprawl,wettergreen1993exploring}. Wheels provide smooth contact and favorable energy scaling~\cite{kim2006isprawl}, but often trade off versatility on unstructured terrain. Hybrid systems such as wheeled-legged robots aim to bridge this gap~\cite{bellicoso2018dynamic,bjelonic2019keep}, but most existing designs rely on actively driven wheels, adding leg inertia and thus reducing energy efficiency in motions that involve significant leg movements.

\begin{figure}
    \centering
    \includegraphics[width=\linewidth, trim=0 1.7cm 0 1.7cm, clip]{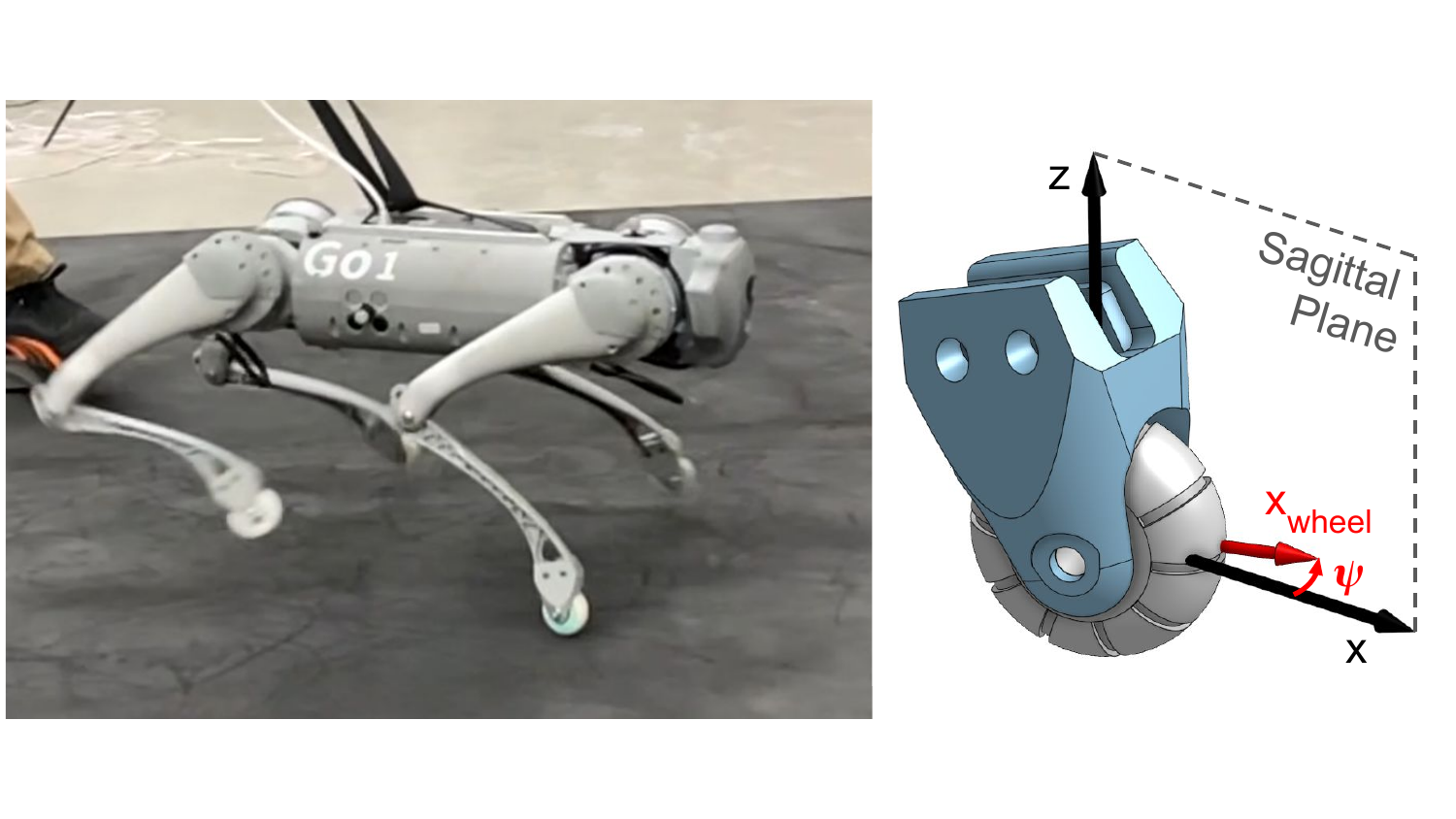}
    \caption{Quadrupedal skating setup with passive wheels. Each foot is equipped with a 3D-printed roller support that holds a passive wheel. In the default standing configuration, the wheel yaw installation angle $\psi$ is defined as the deviation of the wheel’s x-axis, $x_\text{wheel}$, from the x-axis of the robot’s sagittal plane. This angle is our key design parameter.}
    \label{fig:mechanical_design}
\end{figure}

We explore an alternative hybrid mode---\textit{quadrupedal skating with passive wheels}. Equipping each foot with a passive wheel allows a quadruped to retain the stability and maneuverability of legged locomotion while gaining the efficiency and speed of wheeled motion, without motorizing the feet. Figure~\ref{fig:mechanical_design} shows our implementation on a Unitree Go1 using custom 3D-printed roller supports with an adjustable wheel yaw installation angle $\psi$.

Achieving effective skating, however, is challenging. Skating dynamics couple robot posture and velocity under nonholonomic rolling constraints, requiring coordinated regulation of contact forces, body posture, and momentum exchange. With passive wheels, the robot cannot directly command wheel traction. Instead, skating performance depends strongly on the interaction between mechanical design and the learned motor policy. This tight design-control coupling makes \textit{co-design} essential: poor hardware configurations can restrict feasible behaviors, while suboptimal policies may fail to exploit an otherwise capable design.

To address this coupling, we propose a bilevel co-design framework that jointly optimizes mechanical parameters and a control policy. The upper level uses Bayesian Optimization (BO)~\cite{brochu2010tutorial} to efficiently search the design space, while the lower level uses Reinforcement Learning (RL)~\cite{tan2018sim,peng2017deeploco} to train a policy for each candidate design. This BO-RL loop discovers design-policy pairs that enable both efficient and versatile skating.

Our main contributions are as follows:
\begin{itemize}
    \item \textbf{A novel locomotion mode:} We introduce quadrupedal skating with passive wheels, demonstrating its potential for combining speed, efficiency, and maneuverability.
    \item \textbf{A bilevel co-design framework:} We present a BO-RL approach that jointly optimizes wheel installation parameters and the motor control policy.
    \item \textbf{Versatile locomotion demonstrations:} We validate the method on a state-of-the-art quadruped and demonstrate behaviors such as \textit{hockey stop} (rapid braking by turning sideways to maximize friction) and \textit{self-aligning} motion (automatic reorientation to maximize energy efficiency in the moving direction).
    \item \textbf{A systematic study of quadrupedal skating:} We provide, to the best of our knowledge, the first system-level study of dynamic skating motion on quadrupedal robots.
\end{itemize}

Overall, our results suggest that co-designing hardware and control is a practical route to reliable quadrupedal skating behaviors that are difficult to achieve with human-engineered configurations.

%% file: sections/02_related.tex
\section{Related Work}

\subsection{Quadrupedal Locomotion}
Quadrupedal robotics has progressed rapidly, with platforms demonstrating dynamic mobility and robust terrain traversal. Early hydraulic systems such as BigDog established rough-terrain feasibility~\cite{raibert2008bigdog}. Subsequent robots such as MIT Cheetah~\cite{seok2013design} highlighted the effectiveness of model-based control, including whole-body dynamics and model predictive control, for agile running and jumping~\cite{di2018dynamic}. More recently, Reinforcement Learning (RL) has produced highly dynamic locomotion skills and controllers that can match or exceed expert-designed strategies~\cite{hwangbo2019learning,tan2018sim,bao2404deep,ha2025learning}, enabling robust deployment on systems such as ANYmal~\cite{hutter2016anymal}. Despite these advances, pure legged locomotion is typically slower and less energy-efficient than wheeled motion, motivating wheeled--legged hybrids that combine both modalities~\cite{cui2021learning,li2025ctbc,gim2024ringbot,bjelonic2022survey}. However, many of such platforms use actuated wheels, which increase leg inertia and reduce energy efficiency for dynamic motions that involve high speed leg movements.

\subsection{Robot Skating with Passive Wheels}
Skating locomotion with passive wheels offers an appealing way to achieve energy-efficient maneuvers on smooth terrain. Bjelonic et al.\ demonstrated quadrupedal skating on ANYmal using passive wheels and force control to generate glide-and-push sequences, reporting a reduced cost of transport relative to trotting~\cite{bjelonic2018skating}. Chen et al.\ drew inspiration from human skating to synthesize fast and efficient quadrupedal gaits~\cite{chen2020roller}. Ziv et al.\ extended passive wheel skating to a bipedal robot with in-line skates~\cite{ziv2010motion}. Although effective, these approaches rely on heuristics for mechanical design and gait synthesis. In contrast, our paper adopts a hardware-control co-design framework to automatically discover efficient and versatile gaits and the wheel designs that realize them.

\subsection{Robot Hardware and Control Policy Co-Design}
Co-design of hardware and control is a powerful paradigm for unlocking new robot capabilities. Prior works can be broadly grouped into three strategies.

\textbf{Universal policy methods} decouple design and control optimizations by training a single policy across a distribution of designs, which can then serve as a fast evaluator. Examples include design randomization for general locomotion~\cite{feng2023genloco,won2019learning,liu2025locoformer} and design-conditioned policies for actuator optimization~\cite{bjelonic2023learning,belmonte2022meta}. These approaches can be versatile, but often require substantial initial training to cover many suboptimal designs.

\textbf{Joint optimization methods} embed design parameters directly into the RL loop, updating design and control simultaneously. Such approaches can discover novel designs~\cite{schaff2019jointly,ha2019reinforcement}, but may suffer from instability due to the non-stationarity induced by changing body parameters and limited simulator support for online design changes.

\textbf{Bilevel optimization methods} treat co-design as a nested problem: an outer-loop optimizer proposes designs, and an inner-loop learner trains a specialized policy for each candidate. This paradigm has been applied to legged robots and manipulators~\cite{chen2023deep,kim2021mo}, producing expert policies tailored to individual designs. Because repeated RL training is expensive, sample efficiency in the outer loop is crucial; Bayesian Optimization (BO) is well suited for expensive black-box objectives~\cite{shahriari2015taking,brochu2010tutorial}.

Our work adopts this bilevel BO-RL framework and applies it to quadrupedal skating with passive wheels, aiming to jointly discover wheel installation parameters and control policies that yield efficient and versatile behaviors.

%% file: sections/03_quad_skating_via_codesign.tex
\section{Quadrupedal Skating via Co-Design}
Skating with passive wheels requires both mechanical adaptation and tailored control.
Neither component alone is sufficient: wheel orientation shapes which skating gaits are physically feasible, while the learned policy determines whether the robot can stably and efficiently exploit a given design.
This motivates a co-design formulation in which hardware and control are optimized jointly.
In this section, we formalize the BO-RL bilevel co-design problem, describe the roller support design space, and highlight how quadruped leg kinematics constrain the feasible wheel installation angle.

\subsection{Co-Design via Bilevel Optimization}
\begin{figure}
    \centering
    \includegraphics[width=\linewidth, trim=1.5cm 0 1.5cm 0, clip]{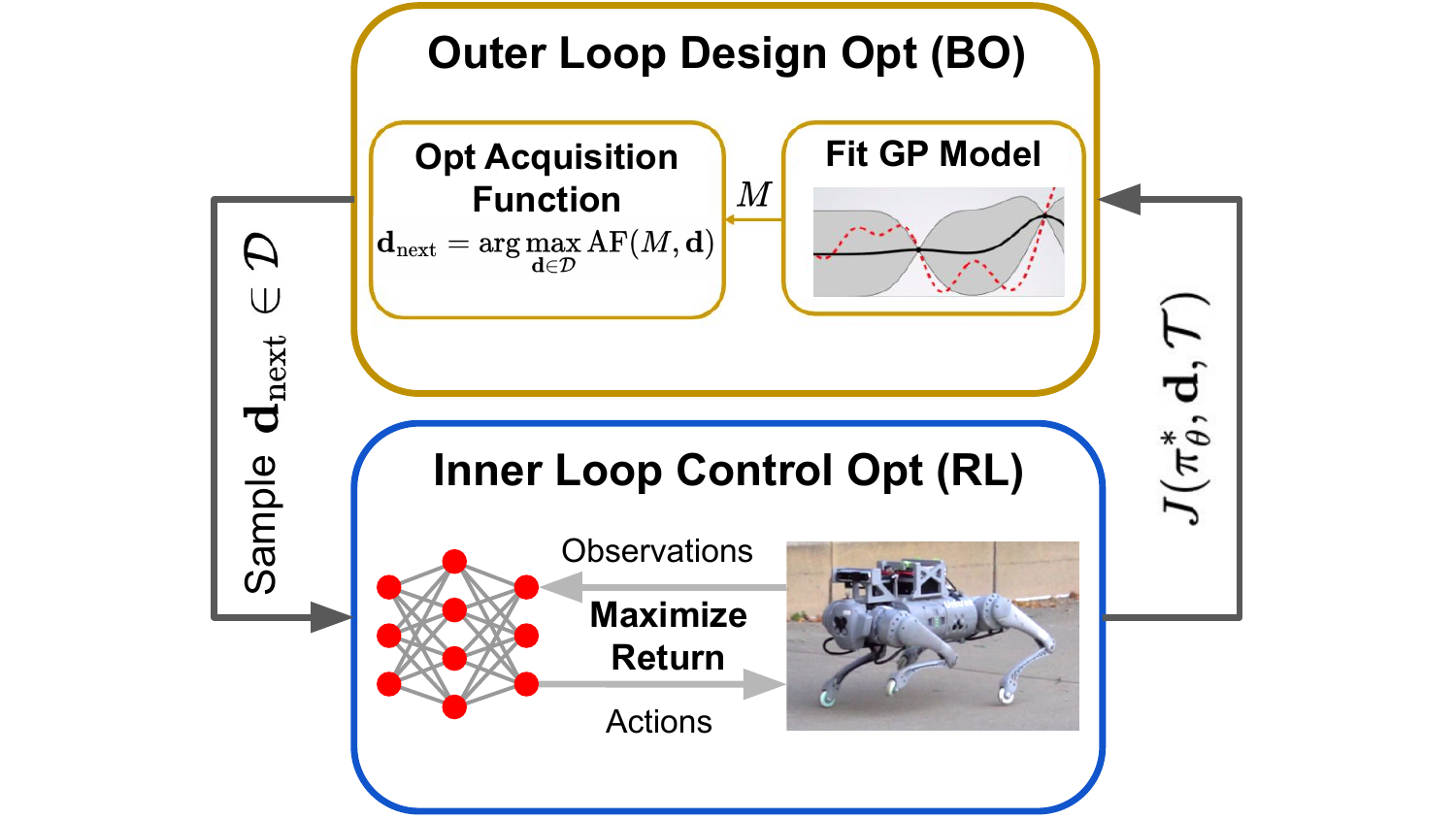}
    \caption{Bilevel co-design framework for quadrupedal skating.
    The outer loop uses BO to propose the next candidate design $\bd_{\txtnext}$.
    For each candidate design, the inner loop uses RL to optimize the policy $\pi_\theta$.
    The resulting performance $J(\pi_\theta^*, \bd, \cT)$ is fed back to BO, enabling efficient search over design-policy pairs that maximize skating performance.}
    \label{fig:framework}
\end{figure}

We formulate quadrupedal skating as the bilevel optimization problem in \Fig{\ref{fig:framework}}, with an inner loop for control policy learning and an outer loop for hardware design.
For a fixed design $\bd$ and a task set $\cT$, the inner loop seeks an optimal policy $\pi_\theta^*(\bd, \cT)$ (parameterized by $\theta$) that maximizes expected discounted return over trajectories induced by executing tasks sampled from $\cT$:
\begin{equation}
    \pi_\theta^*(\bd, \cT) = \arg\max_{\pi_\theta} \,
    \mathbb{E}_{\tau \sim p(\tau \mid \pi_\theta, \bd, \cT)}
    \left[ \sum_{t=0}^{T} \gamma^t r_t \right],
    \label{eq:inner_loop}
\end{equation}
where $\tau$ denotes trajectories generated by controlling the robot with design $\bd$ using policy $\pi_\theta$, $r_t$ is the reward at time step $t$, and $\gamma$ is the discount factor.
We solve the inner-loop problem with RL.
Compared with model-based approaches, RL is well suited to skating because the non-holonomic wheel-ground interactions are handled directly by the simulator, and complex gaits need not be manually specified.

Given the optimized policy, we evaluate performance using a scalar objective $J(\pi_\theta^*, \bd, \cT)$.
The outer loop then searches for the best design within the design space $\mathcal{D}$:
\begin{equation}
    \bd^* = \arg\min_{\bd \in \mathcal{D}} \; J(\pi_\theta^*, \bd, \cT).
    \label{eq:outer_loop}
\end{equation}

Because evaluating $J(\pi_\theta^*, \bd, \cT)$ requires training a full policy and is non-differentiable due to stochastic RL and contact dynamics, the outer loop is naturally treated as black-box optimization.
We therefore use BO in the outer loop, which iteratively fits a Gaussain Process (GP) model $M$ to predict the mean and uncertainty of $-J(\pi_\theta^*,\bd,\cT)$, and maximizes an acquisition function $\text{AF}(M,\bd)$ to select the next candidate $\bd_\txtnext$ while balancing exploration and exploitation. This bilevel formulation enables systematic discovery of design-policy pairs that yield robust and efficient skating.

\subsection{Mechanical Design of Roller Support}
\label{sec:mech_design}
A key component of our platform is a skating foot that mounts a passive wheel at the foot tip of each leg. We replace the robot toes with lightweight 3D-printed supports that connect the legs to the wheels. Our main design parameter is the yaw angle $\psi$, which determines each wheel's rolling direction (Fig.~\ref{fig:mechanical_design}).

\begin{figure}
    \centering
    \includegraphics[width=0.8\linewidth]{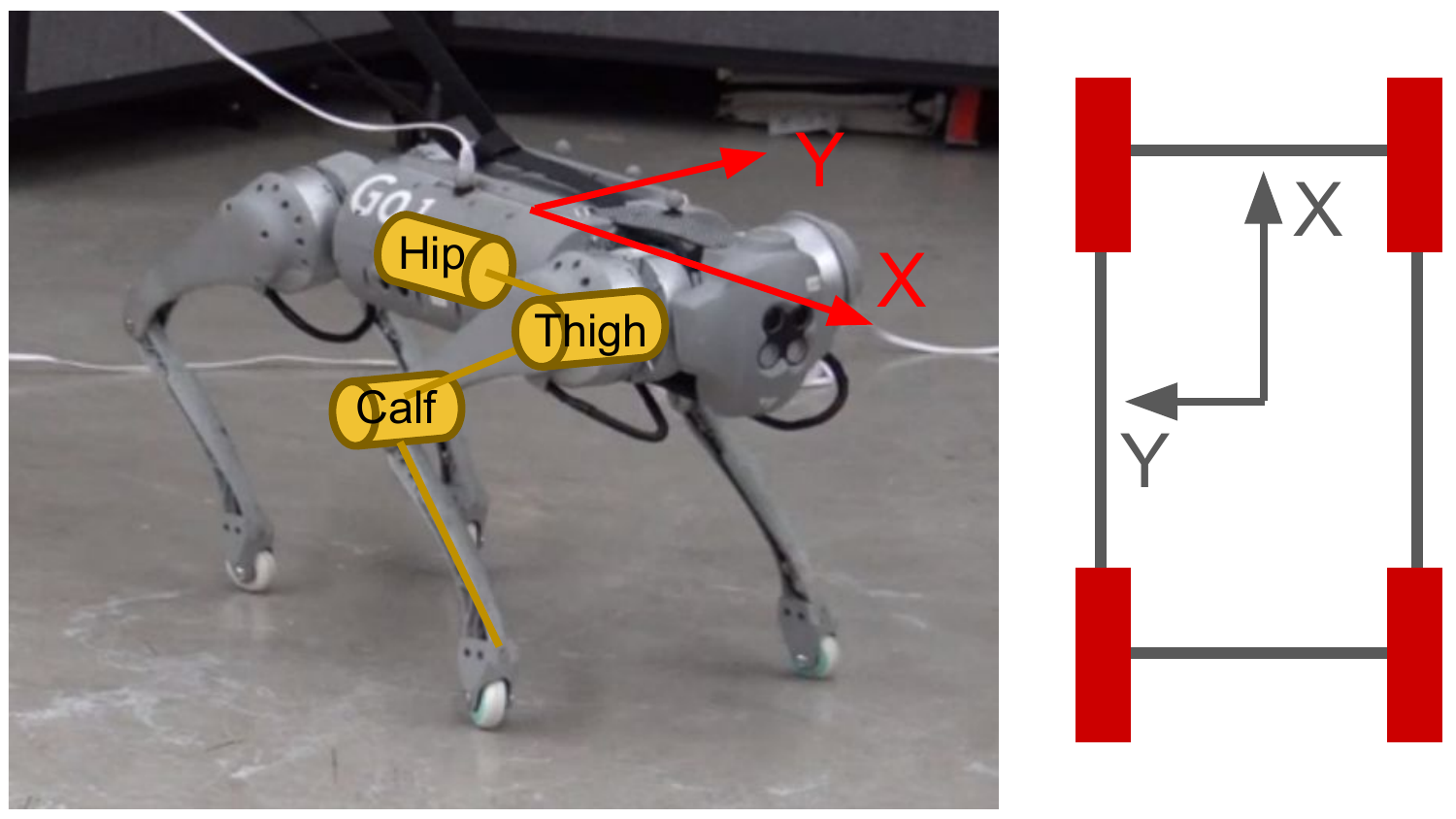}
    \caption{Naive P configuration ($\psi = 0^\circ$) on Unitree Go1~\cite{UnitreeGo1} yields uncontrollable forward velocity $v_x$ due to leg kinematic constraints.
    Similar limitations arise on many state-of-the-art quadrupedal robots.}
    \label{fig:design__PCfg}
\end{figure}

A simple but naive design is the \emph{Parallel Configuration} (P configuration) illustrated in Fig.~\ref{fig:design__PCfg}, where all wheels are aligned with the body x-axis ($\psi = 0^\circ$). Although this appears ideal for forward skating, the leg kinematics of most state of the art quadrupedal robots--including MIT MiniCheetah~\cite{katz2019mini} and Unitree Go1~\cite{UnitreeGo1}, Go2~\cite{UnitreeGo2}, and B2~\cite{UnitreeB2}--constrains the wheel rolling direction to always align with the x-axis when $\psi = 0^\circ$.
As a result, the available traction force along the x direction is almost zero due to negligible rolling friction, making $v_x$ effectively uncontrollable. Consequently, at least one wheel must have a nonzero yaw angle, making the optimal design non-trivial and motivating a systematic co-design approach.

%% file: sections/04_algo_impl.tex



\section{Algorithmic Implementation}

\subsection{Inner Loop Control Policy Optimization via RL}
\block{RL formulation}
\label{sec:rl_formulation}
For each candidate design, we train a control policy with RL in \textit{IsaacLab}~\cite{mittal2025isaac}. We formulate the learning problem as a Partially Observable Markov Decision Process (POMDP) defined by the tuple ($\mathcal{S}$, $\mathcal{O}$, $\mathcal{A}$, $p$, $r$), where $\mathcal{S}$, $\mathcal{O}$, and $\mathcal{A}$ denote the state, observation, and action spaces, respectively, $p(s_{t+1}\mid s_t,a_t)$ is the state transition probability, and $r$ is the reward.

We seek a policy $\pi_\theta(a_t\mid o_t)$ parameterized by $\theta$ that maps observations $o_t\in\mathcal{O}$ to a distribution over actions $a_t\in\mathcal{A}$ by maximizing the expected discounted return:
\begin{equation}
    \pi_\theta^* = \arg\max_\theta \mathbb{E} \left[ \sum_{t=0}^{\infty} \gamma^t r(s_t, a_t) \right],
\end{equation}
where $\gamma \in (0,1)$ is the discount factor. We train $\pi_\theta$ using Proximal Policy Optimization (PPO)~\cite{schulman2017proximal}. The observation space, action space, and reward terms are summarized below.

\block{Observation Space}
\label{sec:obsrv_space}
\Table{\ref{table:obsrv_space}} lists the observations provided to the policy: commanded and measured base linear/angular velocities, joint positions/velocities, and the previous action. The base orientation is represented by projected gravity ${}^{\cB}\bu_g = \bR_{\cW\cB}[0,0,-1]^T$, where $\bR_{\cW\cB}$ is the rotation of the body frame $\cB$ with respect to the world frame $\cW$.

A roller-skating-specific detail is how linear velocity commands are specified and, consequently, how they appear in the observation. \eqref{eqn:lin_vel_obsrv} gives the observation of the commanded linear velocity ${}^{\cB}\bv_{xy}^{\text{obs}}$ (always expressed in the base frame), which can originate from either a base-frame command ${}^{\cB}\bv_{xy}^\cmd$ or a world-frame command ${}^{\cW}\bv_{xy}^\cmd$ that is transformed into the base frame. These two command conventions lead to different tracking rewards. We refer to the two conventions as \emph{Base Frame Command} and \emph{World Frame Command}:
\begin{equation}
    {}^{\cB}\bv_{xy}^\text{obs} = \left\{
        \begin{matrix}
            {}^{\cB}\bv_{xy}^\cmd & \text{(Base Frame Command)} \\
            \bR_{\cW\cB}^T {}^{\cW}\bv_{xy}^\cmd & \text{(World Frame Command)}
        \end{matrix}
    \right.. 
    \label{eqn:lin_vel_obsrv}
\end{equation}
We compare these two formulations in \Sec{\ref{sec:rew_eng_wo_codesign}}.

\block{Action Space}
\label{sec:action_space}
The policy outputs actions $\ba$, which are scaled and biased to obtain the desired joint positions $\bq_j^\cmd$. These targets are tracked by a low-level joint PD controller with gains $k_p$ and $k_d$ (and zero desired joint velocity), causing the joint torques $\btau$ to be
\begin{equation}
    \btau = k_p(\bq_j^\cmd - \bq_j) - k_d\bdq_j.
\end{equation}

\block{Rewards}
\label{sec:rewards}
The reward comprises (i) tracking terms for the base velocity, height, and orientation, and (ii) regularization terms that smooth actions and joint accelerations. To discourage overstretched legs, we penalize the horizontal component of the virtual-leg vector (the vector $\bp_{\text{virt leg}}$ from hip to wheel) via $\LtwoNormSquare{\bp_{\text{virt leg, xy}}}$. We also penalize large collision forces $\boldf$ using $\sum \mathbf{1}_{\{\LtwoNorm{\boldf} > f_0\}}$ where $\mathbf{1}_{\{\cdot\}}$ is one when $\{\cdot\}$ is true and zero otherwise. We also discourage joint position violations $\bq_{j,\text{exceed}}$ by penalizing $\sum \LtwoNormSquare{\bq_{j,\text{exceed}}}$.
\begin{table}
    \centering
    \caption{Observation space.}
    \begin{tabular}{|c|c|}
        \hline
        Name & Expression \\
        \hline
        Base Velocity & ${}^{\cB}\bv$, ${}^{\cB}\bm{\omega}$ \\
        Commanded Base Velocity & ${}^{\cB}\bv_{xy}^d, {}^{\cB}\omega_z^d$ \\
        Projected Gravity & ${}^{\cB}\bu_g$ \\
        Joint Position and Velocity & $\bq_j, \bdq_j$ \\
        Last Actions & $\ba_{\text{prev}}$ \\
        \hline
    \end{tabular}
    \label{table:obsrv_space}
\end{table}
\begin{table}
    \centering
    \caption{Reward terms.}
    \begin{tabular}{|c|c|}
        \hline
        Name & Expression \\
        \hline
        Base Linear Velocity xy & $r_{\bv_{xy}}$ using \eqref{eqn:base_lin_vel_rwd} or \eqref{eqn:world_lin_vel_rwd} \\
        Base Angular Velocity z & $r_{\omega_z}$ using \eqref{eqn:base_ang_vel_rwd} or \eqref{eqn:prioritize_lin_vel_rwd} \\
        Base Height & $(z - z^d)^2$ \\
        Base Orientation & $\LtwoNormSquare{{}^{\cB}\bu_{g,xy}}$ \\
        Base Linear Velocity z & ${}^{\cB}v_z^2$ \\
        Base Angular Velocity xy & $\LtwoNormSquare{{}^{\cB}\bomg_{xy}}$ \\
        Action Rate Minimization & $\LtwoNormSquare{\ba - \ba_{\text{prev}}}$ \\
        Joint Acceleration Minimization & $\LtwoNormSquare{\ddot{\bq_j}}$ \\
        Vertical Virtual Leg & $\LtwoNormSquare{\bp_{\text{virt leg, xy}}}$ \\
        Collision Avoidance & $\sum \mathbf{1}_{\{\LtwoNorm{\boldf} > f_0\}}$ \\
        Joint Position Limit & $\sum \LtwoNormSquare{\bq_{j,\text{exceed}}}$ \\
        \hline
    \end{tabular}
    \label{table:rewards}
\end{table}

Most reward terms are shared across experiments, but the velocity-tracking terms depend on whether commands are specified in the base frame or world frame. For \emph{Base Frame Command}, we use the standard base frame tracking rewards
\begin{subequations}
    \begin{align}
        r_{\bv_{xy}} &= r_{\exp}\left(\LtwoNorm{{}^{\cB}\bv_{xy} - {}^{\cB}\bv_{xy}^\cmd}\right),
        \label{eqn:base_lin_vel_rwd} \\
        r_{\omega_z} &= r_{\exp}({}^{\cB}\omega_z - {}^{\cB}\omega_z^\cmd),
        \label{eqn:base_ang_vel_rwd}
    \end{align}
\end{subequations}
where $r_{\exp}(e) = \exp(-e^2/\sigma)$ denotes an exponential tracking reward for the scalar error $e$.

\begin{figure}
    \centering
    \includegraphics[width=0.8\linewidth]{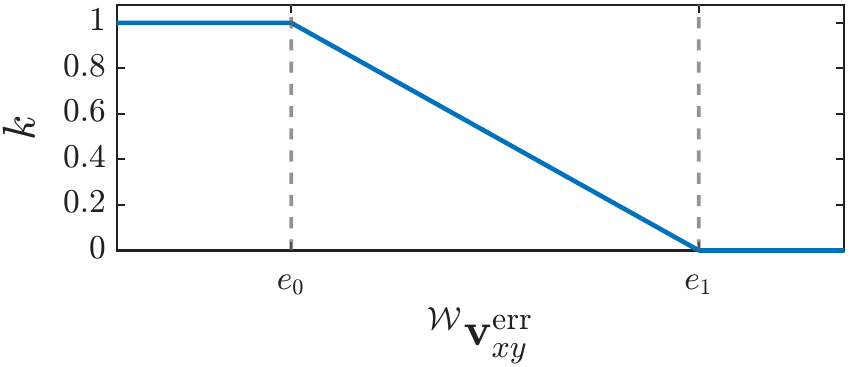}
    \caption{Illustration of \emph{World Frame Command} angular velocity tracking reward scaling to prioritize linear velocity tracking.}
    \label{fig:lin_vel_prioritization}
\end{figure}

For \emph{World Frame Command}, we track linear velocity in the world frame and down-weight yaw-rate tracking when the linear velocity tracking error is large:
\begin{subequations}
    \begin{align}
        r_{\bv_{xy}} &= r_{\exp}\left(\LtwoNorm{{}^{\cW}\bv_{xy} - {}^{\cW}\bv_{xy}^\cmd}\right),
        \label{eqn:world_lin_vel_rwd} \\
        r_{\omega_z} &= k\; r_{\exp}({}^{\cB}\omega_z - {}^{\cB}\omega_z^\cmd), \\
        k &= \left\{\begin{array}{ll}
            1 & ({}^{\cW}v_{xy}^{\err} \leq e_0) \\
            \frac{e_1 - v_{xy,\err}}{e_1-e_0} & ( e_0 < {}^{\cW}v_{xy}^{\err} \leq e_1) \\
            0 & ({}^{\cW}v_{xy}^{\err} > e_1)
        \end{array}\right.,
        \label{eqn:prioritize_lin_vel_rwd}
    \end{align}
\end{subequations}
where ${}^{\cW}v_{xy}^{\err} := \LtwoNorm{{}^{\cW}\bv_{xy} - {}^{\cW}\bv_{xy}^\cmd}$ is the world-frame linear velocity tracking error and $e_0,e_1$ are small/large error thresholds. \Fig{\ref{fig:lin_vel_prioritization}} illustrates this prioritization. It enables strategic body turning, which is detailed in \Sec{\ref{sec:rew_eng_wo_codesign}}.

\subsection{Outer Loop Design Optimization via BO}
\label{sec:outer_loop_bo}
\block{BO Formulation}
Given metric evaluations $J(\pi_\theta^*,\bd,\cT)$ from previously tested designs, each outer-loop BO iteration performs two steps: (i) fit a Gaussian Process (GP) surrogate model $M$ to predict the mean and uncertainty of $-J(\pi_\theta^*,\bd,\cT)$, and (ii) maximize an acquisition function $\text{AF}(M,\bd)$ to select the next candidate $\bd_\txtnext$ while balancing exploration and exploitation. We use a phased acquisition schedule: we begin with Upper Confidence Bound (UCB) and a high exploration coefficient to cover the search space, anneal the exploration coefficient to focus on promising regions, and finally switch to Expected Improvement (EI) to refine the best candidates and improve convergence.

\block{Design Space}
The most general parameterization assigns an independent wheel yaw angle to each leg,
$\mathbf{d} = [\psi_{FR}, \psi_{FL}, \psi_{RR}, \psi_{RL}]$, where FR, FL, RR, and RL denote front-right, front-left, rear-right, and rear-left, respectively. Exploiting the left--right symmetry yields $\psi_{FR}=-\psi_{FL}$ and $\psi_{RR}=-\psi_{RL}$, reducing the search to a 2D space parameterized by $\psi_{\text{front}}:=\psi_{FL}$ and $\psi_{\text{rear}}:=\psi_{RL}$. If the front--rear asymmetry is negligible, we further couple all legs as $[\psi_{FR}, \psi_{FL}, \psi_{RR}, \psi_{RL}] = [-\psi, \psi, \psi, -\psi]$, as illustrated in \Fig{\ref{fig:design__xcfg_1d}}. We report 1D results in \Sec{\ref{sec:baseline_human_designed_roller_skating}}--\Sec{\ref{sec:necessity_of_codesign_for_energy_efficiency_improvement}} and 2D results in \Sec{\ref{sec:combine_rwd_eng_with_codesign}}.

\block{Design Metric}
To assess a design $\bd$ over commands sampled from $\cT$ and to average the stochasticity of the RL, we define an instantaneous metric $f(s,a,c)$ as a function of state $s$, action $a$ and command $c$ in a single time step. Using the rollouts generated during PPO training in IsaacLab, we estimate the design objective $J(\pi_\theta^*,\bd,\cT)$ by Monte Carlo averaging:
\begin{equation}
    J(\pi_\theta^*, \bd, \cT) =
    \frac{1}{N_{\text{step}} N_{\text{env}}}
    \sum_{k=1}^{N_{\text{step}}} \sum_{i=1}^{N_{\text{env}}}
    f(s_{i,k}, a_{i,k}, c_{i,k}),
    \label{eq:mc_estimator}
\end{equation}
which averages over $N_{\text{env}}$ parallel environments and $N_{\text{step}}$ time steps. Commands are resampled from $\cT$ multiple times within the $N_{\text{step}}$-step window. The simulator time step is $0.02\,\text{s}$.

In practice, we aggregate metrics over windows of $N_{\text{step}}=1000$ steps ($\approx 20\,\text{s}$) across $N_{\text{env}}=4096$ parallel environments. Velocity commands are resampled every 200 steps, and environments are reset asynchronously. Although PPO updates occur every 24 steps ($\approx 0.48\,\text{s}$), averaging metrics over 1000-step windows yields more stable estimates.

We instantiate $f$ as an energy-efficiency measure based on the cost of transport (CoT). CoT is computed as the squared joint torque (proportional to motor electrical power) normalized by gravitational force and the $\ell_2$ norm of the actual planar twist $\bxi=[v_x, v_y, \omega_z]$:
\begin{equation}
    \text{CoT} = \frac{\LtwoNormSquare{\btau}}{m g \,\LtwoNorm{\bxi}},
    \label{eq:COT}
\end{equation}
where $\btau$ is the joint-torque vector, $m$ is the mass of the robot, and $g$ is the gravitational acceleration. In the Body Frame Command case, $\bxi = [{}^{\cB}v_x, {}^{\cB}v_y, {}^{\cB}\omega_z]$. In the \emph{World Frame Command} case, $\bxi = [{}^{\cW}v_x, {}^{\cW}v_y, {}^{\cB}\omega_z]$.

%% file: sections/05_results_and_analysis.tex
\section{Results and Analysis}
We present four experiments that evaluate our bilevel BO-RL framework for quadrupedal roller skating. In \Sec{\ref{sec:baseline_human_designed_roller_skating}}, we establish a baseline using a human-engineered wheel design and a control policy trained with \emph{Base Frame Command} tracking. In \Sec{\ref{sec:necessity_of_codesign_for_energy_efficiency_improvement}}, we show that hardware-control co-design is necessary to achieve energy-efficient skating. In \Sec{\ref{sec:rew_eng_wo_codesign}}, we demonstrate transient skating behaviors enabled by a \emph{World Frame Command} tracking reward. Finally, in \Sec{\ref{sec:combine_rwd_eng_with_codesign}}, we combine co-design (\Sec{\ref{sec:necessity_of_codesign_for_energy_efficiency_improvement}}) and reward engineering (\Sec{\ref{sec:rew_eng_wo_codesign}}) to achieve low energy consumption and strong transient velocity tracking. To obtain base linear velocity and orientation estimations on hardware, we use OptiTrack~\cite{optitrack} motion capture system for indoor experiments and OpenVINS~\cite{geneva2020openvins} for outdoor experiments.
\begin{figure}
    \centering
    \includegraphics[width=0.9\linewidth, trim=0 1.2cm 0 1.7cm, clip]{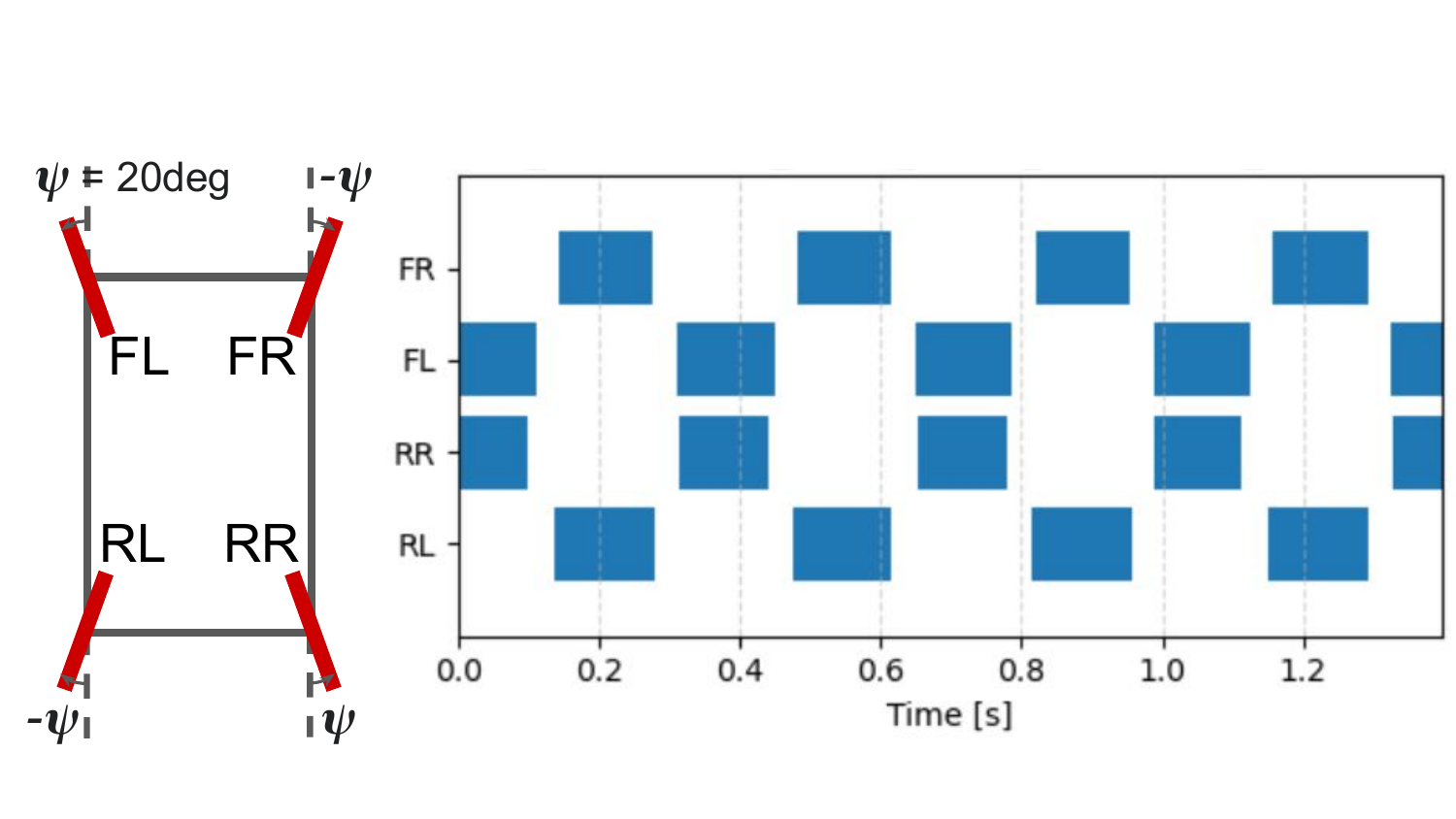}
    \caption{Illustration of the human-engineered roller wheel angles and the resultant gait pattern.}
    \label{fig:design__xcfg_1d}
\end{figure}

\subsection{Baseline via Human-Engineered Roller Skating}
\label{sec:baseline_human_designed_roller_skating}
This section describes the wheel yaw angles of the human-engineered design as the baseline. To address the controllability issue discussed in \Sec{\ref{sec:mech_design}}, we introduce a minimal design modification in which all wheel angles are parameterized by a single non-zero yaw angle $\psi$. As shown in \Fig{\ref{fig:design__xcfg_1d}}, the resulting gait is a trotting gait, where the diagonal legs alternate to contact with the ground. Intuitively, the robot couples each diagonal leg pair to act like a single roller skate.
\begin{figure*}
    \centering
    \includegraphics[width=\textwidth, trim=0 5.3cm 0 5.3cm, clip]{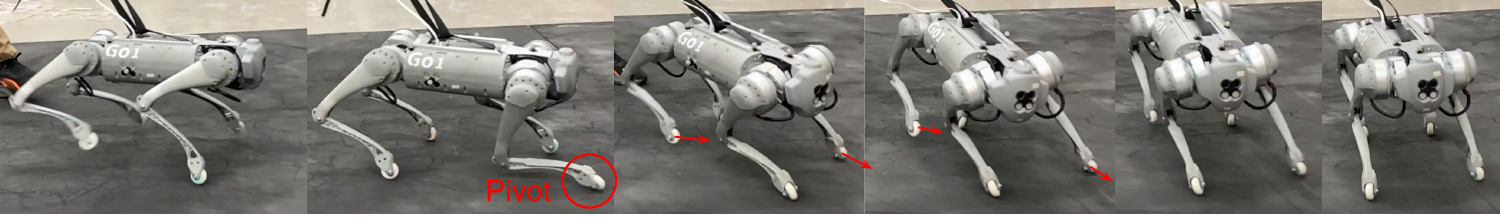}
    \caption{A rapid stopping strategy known as the ``hockey stop'' emerged when using \emph{World Frame Command} with the human-engineered design in \Fig{\ref{fig:design__xcfg_1d}}. When the robot is moving forward quickly and is suddenly commanded to stop, it rotates its body so that its lateral direction (which corresponds to the direction of maximal wheel friction) aligns with the velocity direction, thereby generating a large lateral friction force for deceleration.}
    \label{fig:cool_behavior__hockey_stop}
\end{figure*}
\begin{figure*}
    \centering
    \includegraphics[width=\textwidth, trim=0 3.5cm 0 3.5cm, clip]{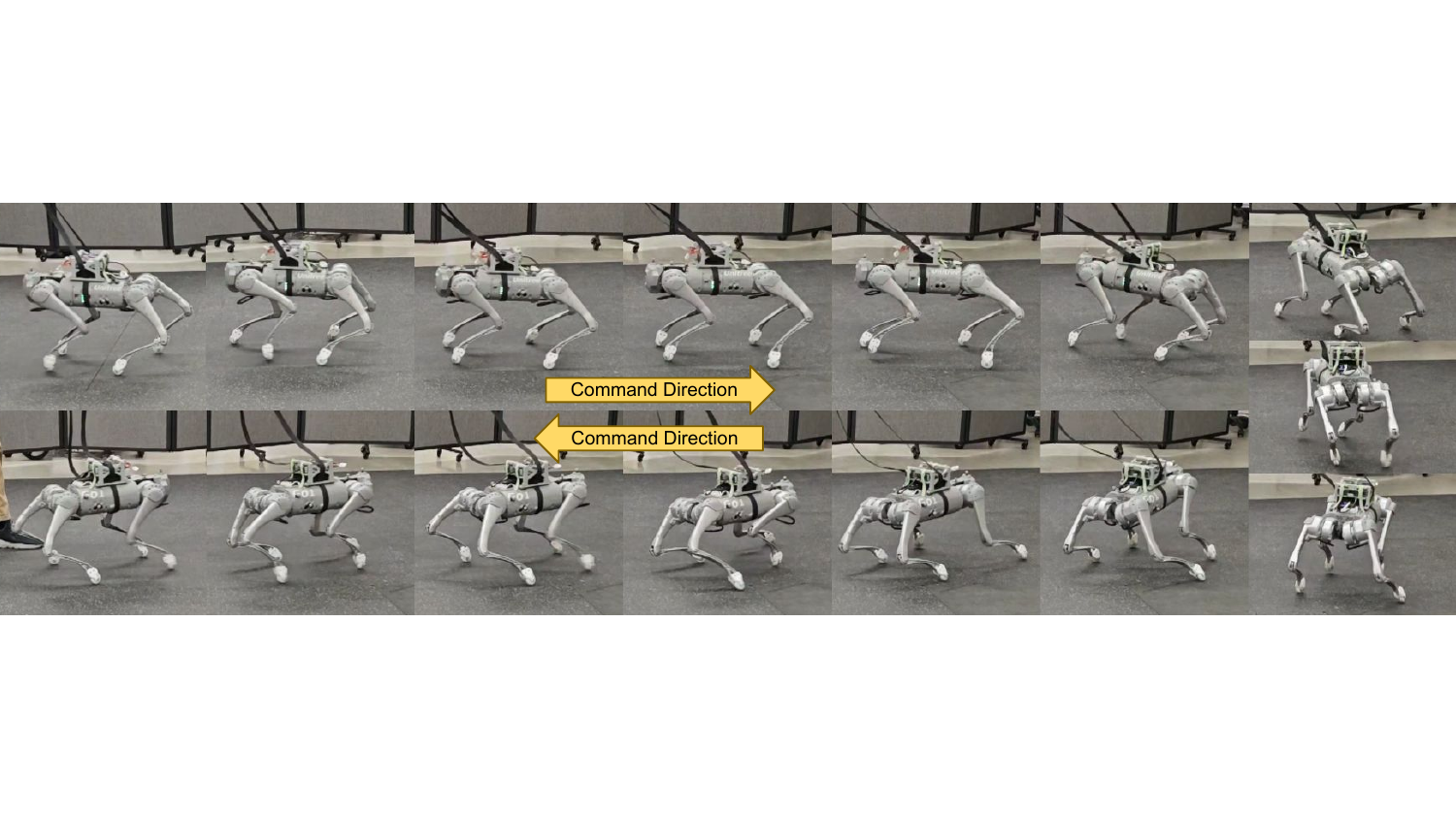}
    \caption{An energy-efficient self-aligning behavior emerged when combining \emph{World Frame Command} with the 2D design parameterization. For the optimal design in \Fig{\ref{fig:design__BO_2D}}, the backward direction is the most energy-efficient. As a result, the robot learns to consistently align its backward direction with the commanded velocity direction.}
    \label{fig:cool_behavior__self_align}
\end{figure*}

\subsection{Necessity of Co-Design for Energy Efficiency}
\label{sec:necessity_of_codesign_for_energy_efficiency_improvement}
We next demonstrate the importance of co-optimizing design and control. In the inner-loop RL used to train control policies, we use \emph{Base Frame Command} observations and rewards, i.e., we specify the velocity command relative to the robot's orientation. This choice enables consistent comparisons of energy efficiency in specific command directions. In the outer-loop BO, we optimize over the minimal 1D design space defined by the yaw angle $\psi$ of the wheels, which generates the four yaw angles of the wheels shown in \Fig{\ref{fig:design__xcfg_1d}}. The BO objective is the average CoT defined in \eqref{eq:mc_estimator}--\eqref{eq:COT}.

\begin{figure}
    \centering
    \includegraphics[width=\linewidth]{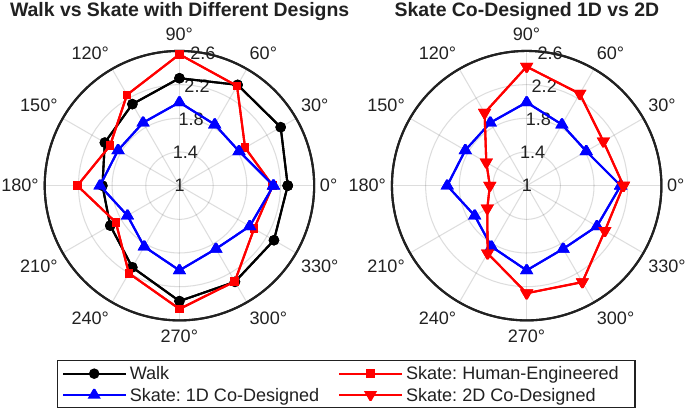}
    \caption{Directional CoT comparison. The left figure compares the CoT of walking and skating using human-engineered and 1D co-designed roller wheels. The right figure compares skating with 1D and 2D co-designed roller wheels.}
    \label{fig:CoT_walk_vs_skate}
\end{figure}

\Fig{\ref{fig:CoT_walk_vs_skate}} reports steady-state CoT in polar coordinates while tracking a body-frame linear velocity command with magnitude 1.5 m/s in different directions. The polar angle is the command direction $\alpha = \arctan2({}^{\cB}v_x^\cmd, {}^{\cB}v_y^\cmd)$, and the radius is the corresponding CoT. The left subfigure shows that the 1D BO-RL co-designed solution is more energy efficient than walking in almost all directions, whereas the human-engineered design is only more efficient in the forward direction ($\alpha = 0\,\text{deg}$). This indicates that co-optimizing design and control is critical for synthesizing energy-efficient quadrupedal roller skating behaviors.

\subsection{Comparison between Base Frame Command and World Frame Command}
\label{sec:rew_eng_wo_codesign}
We now study the effect of tracking linear velocity in the world frame rather than the base frame. With wheel installation angles fixed to the human-engineered design in \Fig{\ref{fig:design__xcfg_1d}}, the robot's forward direction is close to the wheel rolling directions. This is energy efficient, but it provides limited friction authority to change the robot's velocity. In contrast, the sideways direction offers substantial friction for velocity tracking but is less energy efficient because the wheels can barely roll laterally.

\emph{World Frame Command} allows the policy to strategically switch between these regimes to balance energy efficiency and velocity tracking. \Fig{\ref{fig:cool_behavior__hockey_stop}} illustrates a scenario in which the robot moves quickly and is suddenly commanded to stop. During steady-state motion, the policy aligns the robot's forward direction with the velocity direction to maximize efficiency. When commanded to stop, the robot rotates to align its sideways direction with the velocity direction, producing a large lateral friction force for deceleration. The resulting behavior resembles a pivoting maneuver: the front-right wheel acts as a pivot, while the front-left and rear-right wheels continue rolling, effectively rotating the body about the pivot wheel. The robot then exploits sliding friction from all wheels to achieve a rapid stop.

This maneuver is known in roller skating as the ``hockey stop'' and is among the quickest ways to stop. It cannot be learned with \emph{Base Frame Command} because base-frame velocity tracking implicitly constrains the body's orientation relative to the commanded velocity, preventing the policy from reorienting to exploit lateral friction for braking. With \emph{World Frame Command}, the stop time from an initial forward speed of $2\,\text{m/s}$ is reduced by approximately 50\% compared with the \emph{Base Frame Command} policy. This result highlights how world-frame tracking can induce emergent strategies that select body orientation to maximize control authority.

\subsection{Combining Reward Engineering with Co-Design}
\label{sec:combine_rwd_eng_with_codesign}
\begin{figure}
    \centering
    \includegraphics[width=0.9\linewidth]{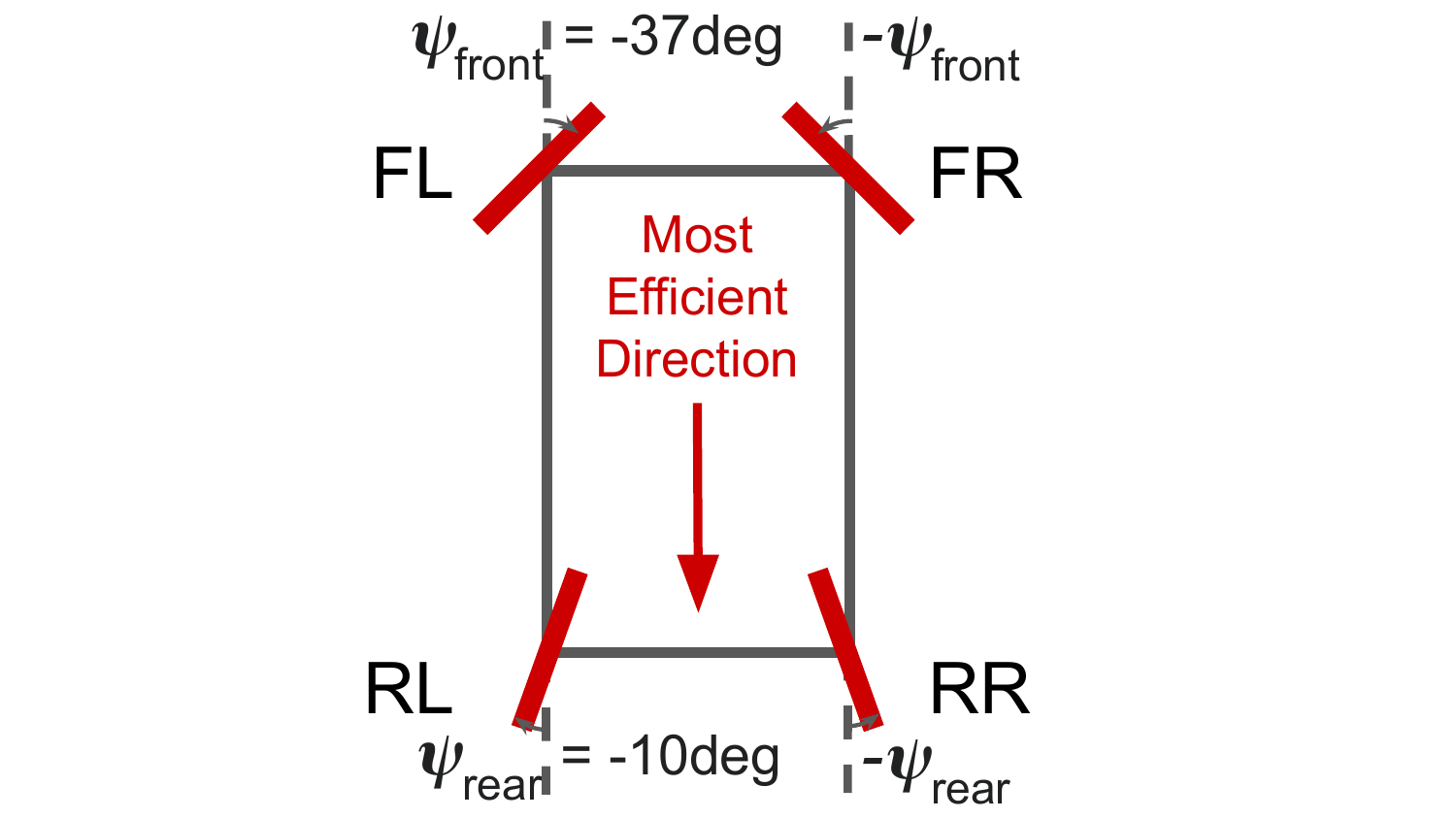}
    \caption{2D co-design optimal design.}
    \label{fig:design__BO_2D}
\end{figure}

Combining \emph{World Frame Command} with the 2D design parameterization in \Sec{\ref{sec:outer_loop_bo}} yields the design shown in \Fig{\ref{fig:design__BO_2D}}. This design exhibits a self-aligning behavior (\Fig{\ref{fig:cool_behavior__self_align}}): without explicitly tracking an angular velocity target, the robot learns to align its backward direction with the commanded linear velocity. This reduces average CoT by 14.6\% relative to the 1D co-design.

To understand this mechanism, we fix the 2D-optimized design and retrain the control policy with \emph{Base Frame Command} to evaluate directional CoT. As shown in the right subfigure of \Fig{\ref{fig:CoT_walk_vs_skate}}, the 2D co-optimized design is less efficient for sideways skating but achieves its lowest CoT in the backward direction---the orientation that the \emph{World Frame Command} policy preferentially aligns with during execution. These results suggest that, when reward engineering encourages strategic body-orientation selection, the wheel installation angles can be optimized for peak efficiency in a preferred direction rather than uniform efficiency across directions, yielding the most energy-efficient skating behavior in this work.
\vspace{-6pt}

%% file: sections/06_conclusion_and_future_work.tex
\section{Conclusion and Future Work}
We presented quadrupedal skating with passive wheels as a new locomotion mode that merges the efficiency of wheeled motion with the agility of legged robots. Through a bilevel BO-RL co-design framework, we identified design–policy pairs that enable efficient skating and emergent behaviors such as hockey stop and self-aligning motion.

Future work will extend this study beyond flat terrain to outdoor and uneven environments, explore lightweight and durable wheel modules, synthesize better reward engineering, and develop more sample-efficient co-design methods. Integrating skating with walking and running modes is another exciting direction toward versatile and multimodal robots.

\balance






